\pdfoutput=1

\documentclass[11pt]{article}

\usepackage{ACL2023}

\usepackage{times}
\usepackage{latexsym}
\usepackage{url}
\usepackage{amsmath}
\usepackage{amssymb}
\usepackage{multirow}
\usepackage{float}
\usepackage{color,soul}
\usepackage{tikz}
\usetikzlibrary{bayesnet}
\usepackage{subcaption}
\usepackage{graphicx}
\usepackage{bbm}
\usepackage{enumitem}

\usepackage[T1]{fontenc}

\usepackage[utf8]{inputenc}

\usepackage{microtype}

\usepackage{inconsolata}

\title{Exploring the Learning Capabilities of Language Models \\ using {\sc LeverWorlds}}

\author{Eitan Wagner$^\dagger$\quad Amir Feder$^\ddagger$\quad Omri Abend$^\dagger$ \\
         $^\dagger$Hebrew University of Jerusalem \quad $^\ddagger$Columbia University\\
         \texttt{eitan.wagner@mail.huji.ac.il}}

\begin{document}
\maketitle

\begin{abstract}
Learning a model of a stochastic setting often involves learning both general structure rules and specific properties of the instance. This paper investigates the interplay between learning the general and the specific in various learning methods, with emphasis on sample efficiency. 
We design a framework called {\sc LeverWorlds}, which allows the generation of simple physics-inspired worlds that follow a similar generative process with different distributions, and their instances can be expressed in natural language. These worlds allow for controlled experiments to assess the sample complexity of different learning methods.
We experiment with classic learning algorithms as well as Transformer language models, both with fine-tuning and In-Context Learning (ICL). Our general finding is that (1) Transformers generally succeed in the task; but (2) they are considerably less sample efficient than classic methods that make stronger assumptions about the structure, such as Maximum Likelihood Estimation and Logistic Regression. This finding is in tension with the recent tendency to use Transformers as general-purpose estimators. 
We propose an approach that leverages the ICL capabilities of contemporary language models to apply simple algorithms for this type of data. Our experiments show that models currently struggle with the task but show promising potential.\footnote{Code is provided at \url{https://github.com/eitanwagner/leverworlds}}
\end{abstract}

\section{Introduction}

Many statistical learning settings combine two types of challenges: discovering the underlying persistent \textit{structure} or representation of the problem, and modeling the context-dependent \textit{variability} \citep{bengio2013representation}. 
These two factors differ in their generality -- the structure is shared by many cases that might differ in their variability. 

For example, assume we want to know how long a typical object will take to reach the ground when dropped from a building in some city. Assuming this knowledge has not been directly reported, we must acquire it from the data. We can conduct experiments by dropping different balls from different buildings. 
Learning involves acquiring two types of knowledge -- one is the physical rules of free fall (e.g., that the mass does not influence the time of the fall), and the other is the distribution of the heights of the buildings in this city (assuming we cannot directly measure the heights). Both types are induced from experiments, but the first is universal, and as such it might already be known based on experiments conducted in a different place. 

Recent Large Language Models (LLMs) are used as general purpose learners, as almost any task that does not require multiple modalities can be formulated as text-to-text or text completion. Therefore, the model must learn both the world and stochastic model for effective learning. 
As typically models are trained with likelihood-based objectives, the models' output confidence for completions should reflect the underlying distribution of the data.

\begin{figure*}[t]
\centering
\includegraphics[trim={1cm 0 0 0}, clip,scale=0.85]{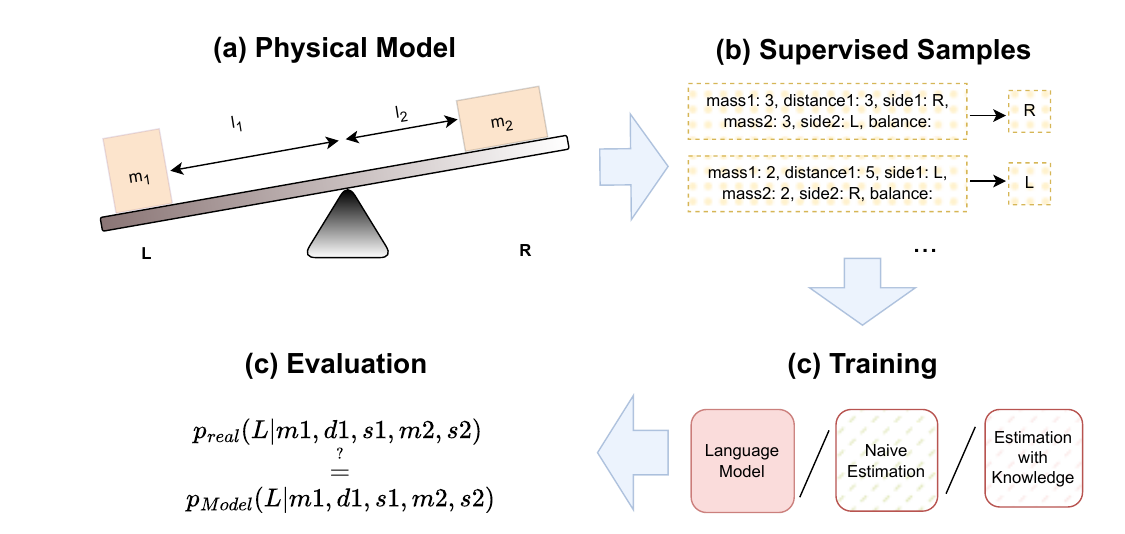}
\caption{Overview of our experiments. First, we generate a physical model, then we sample from the model and train a language model to predict the output. We then evaluate the model's probability estimations.}
\label{fig: overview}
\end{figure*}

The density estimation task from observations is well-studied in statistics and machine learning.
Different methods for parameter estimation differ in their assumptions (or parameterizations). Generally, a model with fewer assumptions can better fit the true unknown) distribution, compared to a model with many assumptions. This is described as having lower \textit{bias}. On the other hand, when a model is more flexible it is more sensitive to noise in the specific training set, hurting generalization. This is described as having higher \textit{variance}. This tradeoff between these two phenomena is known as the \textit{bias-variance tradeoff}. The tradeoff has implications on the \textit{sample complexity} of a model, as high variance might require training sets with more samples.

In this work, we investigate the place of LMs within this tradeoff. To show this, we design a simplistic yet rich world of physical models. We train LMs to learn these worlds from samples (see Figure \ref{fig: overview} for a schematic overview). 
We show that even in this simple setting, although better than extremely na\"{i}ve models, LLMs are substantially less sample efficient than classical statistical estimation methods that have a stronger inductive bias (e.g., logistic regression). Analysis of our experiments shows that this sample inefficiency can be ascribed more to the underlying structure model than to the variability.

We further show that leading off-the-shelf LLMs, like GPT4o\footnote{\url{https://openai.com/index/hello-gpt-4o/}}, fail on this task in an in-context-learning setting. However, inspired by the underlying structure vs. variability distinction, we propose a method for introducing inductive bias into these models by guided construction of a learning pipeline that includes classical models. We find that although challenging for all models, some models show a substantial improvement over others, revealing a promising trend.
This further supports the centrality of the distinction.

Our framework, {\sc LeverWorlds}, is also a contribution in its own right. It has many appealing traits: (1) it is based on real physical laws, and is thus not a ``toy example''; (2) it includes many variations, all of which are simple to learn when the physical rules are known but complicated otherwise; (3) it follows a generative process, which allows for the generation of arbitrarily large sets of supervised samples, with confidence scores, for training and testing.

Our contributions are: (1) we present a framework for experiments in a physical setting;
(2) we show that LMs succeed in learning world models, but they are substantially inferior (in terms of sample complexity) to classical models with stronger assumptions;
(3) we show the distinction between learning the world model and learning the latent model;
(4) we propose methods for using LLMs in combination with classical models, with promising initial results.

\section{Related work}

\subsection{World Models}
\paragraph{In pretrained LLMs.}
Many works evaluate and explore the extent to which pretrained LLMs encode world models.
\citet{abdou-etal-2021-language} show a correspondence between textual color representations in LMs and a perceptually meaningful color space.
\citet{gurnee2024languagemodelsrepresentspace} show that LLMs learn linear representations of space and time across multiple scales.
\citet{vafa2024evaluatingworldmodelimplicit} propose evaluation metrics for world model recovery and show that world model representations are still inconsistent even for models with high accuracy in prediction tasks.

\paragraph{Learning world models from examples.}
Some works demonstrate the capabilities of Transformers in learning artificial domains with deterministic rules.
\citet{demeter-downey-2021-whos} and \citet{toshniwal2022chesstestbedlanguagemodel}
show that Transformers trained on chess games can learn to track pieces and predict legal moves with high accuracy. \citet{demeter-downey-2021-whos} additionally demonstrate the capabilities in baseball game states.
\citet{li2024emergentworldrepresentationsexploring} show that Transformers trained on Othello games can be linearly mapped to the true board.

\citet{liu2024llmslearngoverningprinciples} show that Transformers can successfully learn dynamical systems from in-context sequences alone, thus expanding to probabilistic worlds that describe real events.
\citet{patel2022mapping} show that LLMs have the ability to map descriptions to actions in a grid world, based on in-context examples.

\paragraph{Integrating world models and language.}
\citet{richens2024robustagentslearncausal} show that robustness under distributional shifts requires an approximate causal model of the data generation. 
\citet{chen2023structured} propose methods to incorporate structures, given as Bayesian Networks, into neural networks and apply these methods to tabular and visual data.
\citet{wong2023word} propose a framework that combines neural language models with probabilistic models, enhancing the models' ability to capture and utilize world knowledge effectively. 
\citet{feng2024chessgpt} combine examples and natural language instructions to train a chess model.
\citet{pmlr-v202-carta23a} propose methods for grounding LLMs in interactive textual environments based on online Reinforcement learning.

\subsection{Finetuning and In-Context Learning} 
A common training paradigm in NLP is to divide training into self-supervised pretraining and task-specific supervised finetuning \cite{devlin-etal-2019-bert}.
\citet{brown2020language} showed that large-scale LMs can be used as few show learners, with the task-specific instructions given as a prompt. Performing tasks with prompts only is known as In-Context Learning (ICL) and is gaining popularity \citep{team2023gemini, dong2024surveyincontextlearning}. 

ICL is more memory efficient than finetuning and some works argued that it generalizes better \cite{awadalla-etal-2022-exploring}. Other works showed that for models with similar sizes, finetuning can generalize well or even better than ICL \cite{mosbach-etal-2023-shot}.

Despite the power of ICL, \citet{liu-etal-2024-lost} show that LLMs struggle with long context prompts and degrade significantly when the relevant information is in the middle of a long prompt. 
\citet{min-etal-2022-rethinking} analyze the role of demonstrations in ICL. They find that the gold truth labels have little effect and suggest that ICL may not be appropriate when the input-label correspondence is not already captured in the LM.

\subsection{Bias-variance Tradeoff}
The bias-variance tradeoff is a fundamental concept in machine learning, where a model's capacity to generalize from training data is balanced against its ability to fit the training data accurately. Some recent work, such as \citet{neal2019modern}, shows that neural networks can defy the traditional bias-variance tradeoff with increased width. Similarly, \citet{dar2021farewell} discuss how overparameterization in neural networks can lead to better generalization. 
The tradeoff is important when considering the inductive biases of different models. Parametric models, which assume a specific form for the underlying distribution, often exhibit different bias-variance characteristics compared to nonparametric models, which do not make such assumptions. 

\section{The {\sc LeverWorlds} Environment} 

\begin{figure}[t]
\centering
\includegraphics[width=0.49\textwidth]{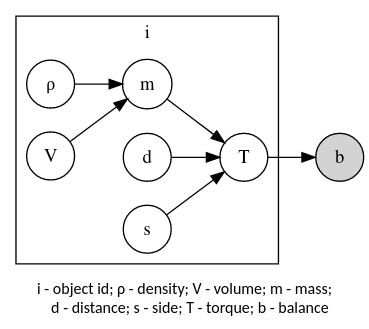}
\caption{Causal graph for balance on a lever. Different worlds differ by the number of objects, by the optional use of density and volume, and by whether the intermediate variables are observed or not.}
\label{fig: causal}
\end{figure}

As empirical validation for our arguments, we experiment with a case in which the world has a simple structure and assess the sample complexity when using language models.
Understanding the underlying world model can significantly reduce the effective sample complexity of the task, allowing accurate estimations with a small amount of training data.

We design a framework for generating worlds that enables efficient sampling and estimation. For each generated world, we design models for estimating the distribution of data based on observed samples. 

\subsection{Setting}

For the general framework, we construct worlds by placing weights on a lever. The lever is placed on a fulcrum with some random number of weights on each side. Each world setting is defined by a causal graph (see Figure \ref{fig: causal}), where different worlds differ by the number of weights (values of $i$), the distribution of the variables, and which variables are latent. 
The variables in the model are density ($\rho$), volume ($V$), mass ($m$), distance from the fulcrum ($d$), and side ($s$). $\rho, V, m$ and $d$ are real numbers and $s$ can be $\pm 1$. The torques ($T$) are determined by the other variables according to the formula $T=s \cdot d \cdot m$. For balance ($b$), $b=1$ if $\sum_i T_i>=0$ and otherwise $b=-1$, corresponding the Left and Right, respectively. Masses are determined by the density and volumes, if not latent, according to the formula $m= \rho \cdot V$.

An input $x$ is a sequence of assignments to all the visible variables. We denote the sequence of assignments of the latent variables by $l$. Given $x$ and $l$ the outcome $b$ is deterministic. 

In each given world, sampling is straightforward -- we follow the graph and sample the outcome, which is Left or Right.
Since the true model is known, we can generate as many samples as we want. This way we can build training sets of arbitrary sizes. During inference, we focus on the output probability given the visible inputs.

We note that each world is defined by two components that must be learned. The first is the general structure of how the outcome depends on a fully observable input, determined by the laws of physics. The second is the case-specific variation which depends on the latent distribution.

The physical model is common to all the settings and is also faithful to the known laws of physics, so models that incorporate general knowledge do not need any samples for this. The latent model is independent between settings and must be learned based on samples in a density estimation process.

\subsection{Evaluation}

\paragraph{Distribution Similarity}
The true distribution $p(y|x)$ is known. A learning model yields an estimator $\hat{p}(y|x)$. Evaluation can be done by comparing $p$ and $\hat{p}$.

The main evaluation is simply the distance between the distributions. We decided to use the expected Total-Variation (TV) distance $E_x [ d_{TV}(\hat{p(y|x)}, p(y|x))] = \frac{1}{2} \cdot E_x [ |\hat{p}- p| ]$. Other measures, like the Jensen-Shannon distance, gave similar empirical results, so we decided to use TV due to the simplicity in deriving concentration bounds.

\paragraph{Structure Similarity}
We also include an evaluation that addresses the dependency structure of a learned model by measuring the effect of minimal input changes on the output. For example, given a pair of inputs that have the same values except for the mass of an object on the left, then the outputted probability for ``L'' must be larger for the input with the larger mass. 

Formally, for a set of inputs $\{x\}^m_{i=1}$, we generate a modified set $\{x^*\}^m_{i=1}$, by randomly choosing one index $j$ in $x_i$ and changing it. 
Denoting the side of the $j$-th index by $s_j$, we define $\Delta x_i := s_j \cdot (x^*_{i_j} - x_{i_j})$ and $\Delta p_i := p(b=L|x^*_i) - p(b=L|x_i)$.
We know that, for the ground truth model, $sign(\Delta x_i) = sign(\Delta p_i)$.
The structure score for a model $M$ is then defined as 

\begin{small}
\begin{multline}
    Score = \frac{1}{m}|\{i | sign(\Delta p^{real}_i) = sign(\Delta p^{M}_i) \}|
\end{multline} \label{structure-score}
\end{small}

\section{Experiments}

In this section, we present the baseline and Transformer models that we use for the experiments. Transformers, as well as some baseline models, are task-agnostic, whereas some baselines leverage task-specific properties.

A different way to describe the difference between models is by the strength of their assumptions about 
the physical setting (but also about the latent variable, since, e.g., the optimal model assumes it is normally distributed).

\subsection{Baseline Methods}

\paragraph{Na\"{i}ve MLE.} 
In this method, we make no assumptions regarding the relationship between inputs. We do, however, assume that we know what the random variables in the input are. Given this, the method independently estimates a Bernoulli distribution for the output, for each possible input, as the frequency of the output in the data. 

Formally, the estimator for each input $x$ is 
\begin{equation}
  \hat{p}(Y=1|x) =
    \begin{cases}
      \frac{N_{x,Y=1}}{N_x} & \text{if } N_x>0\\
      0.5 & \text{otherwise}\\
    \end{cases}       
\end{equation}
where $N_x$ is the number of samples in which the input is $x$ and $N_{x, Y=1}$ is the number of samples with input $x$ and output 1.

\paragraph{Linear Models.}
In another baseline, we perform Logistic Regression (LR) for the output given features of the input. We use polynomial features of degree up to $4$. The model assumes simple relationships between the output and the inputs and their interactions. Also in this method, we assume that we know what the random variables are.
In contrast to the first baseline (which requires estimators for each possible input), the second baseline requires a small number of parameters. 

\paragraph{MLE with knowledge of the full structure.}
In this method, the model knows the underlying rules of the output when all the variables are given. This must include the latent variables which we denote by $L$. Specifically, the function
$$ q(y|x, l) = \mathbbm{1}_{\sum_i T_i > 0} $$
is provided, where $y$ is the output, $x$ is an input, and $l$ is some value of the latent variables.

Learning in this model is simply done by estimation of the distribution of the latent variables.
Formally, the MLE density estimator for the latent variable $l$ at point $c$ is

$$
\hat{p}(l=c)
= \frac{1}{N} \sum_i p(l=c|x_i, y_i)
$$
where the training data is $\{(x_i, y_i)\}^N_{i=1}$.

The estimator for the output is the marginal
$\hat{p}(y|x) = \sum_{c} \hat{p}(l=c) \cdot q(x, l=c)$.\footnote{For simplicity, we assume all variables are discrete. For continuous variables, density should be used instead of mass and the sum should be replaced with an integral.}

\subsection{Transformer Fine-tuning}

Our main investigation addresses the capabilities of general-purpose text models in simple tasks.

\paragraph{Formulation as a text completion task.}
To formulate the task as language completion, we convert the data into text. We list the visible variables by their names with their values. We use the following template:

\begin{quote}
    object1 density: <$v_1$>, object1 volume: <$v_2$>, object1 distance: <$v_3$>, object1 side: <$v_4$>, object1 mass: <$v_5$>, ... balance: <$v_6$>
\end{quote}
where $v_1, v_2, \dots$ represent the corresponding values. The values of the side and balance are given as ``L'' or ``R''.

We train generative language models to predict the outcome by generating ``L'' or ``R'' and we measure the probability of generating each one.

We can add a prompt to give additional information regarding the setting.
However, in our experiments, we found that this prompt has little effect on the performance. We therefore report results without it.

\begin{figure*}[t]
  \centering
  \subfloat[Total-Variation Distance]{%
    \includegraphics[width=0.33\textwidth]{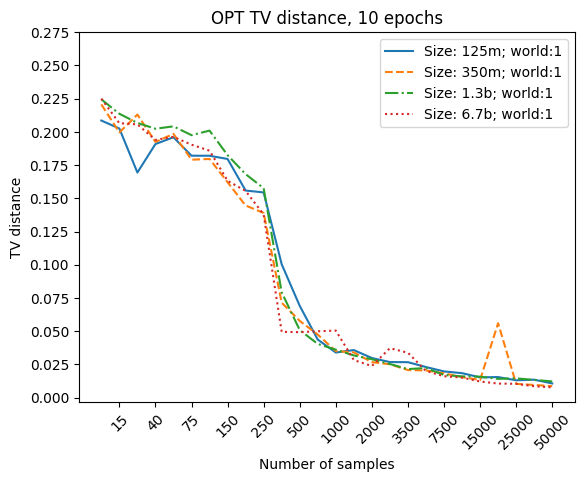}%
    \label{fig:fig1a}%
  }\hfill
  \subfloat[Structure Scores]{%
    \includegraphics[width=0.33\textwidth]{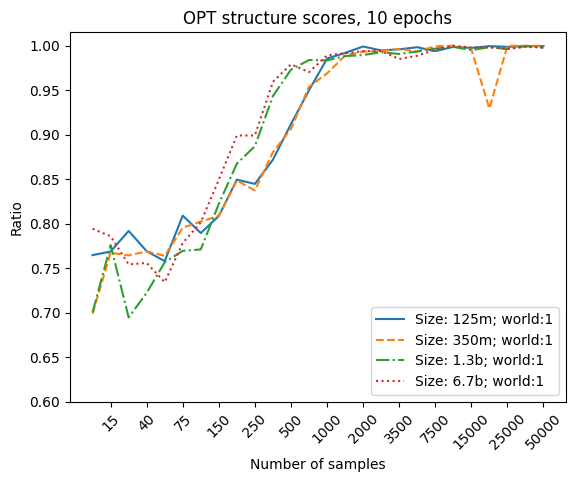}%
    \label{fig:fig2a}%
  }\hfill
  \subfloat[Perplexity]{%
    \includegraphics[width=0.33\textwidth]{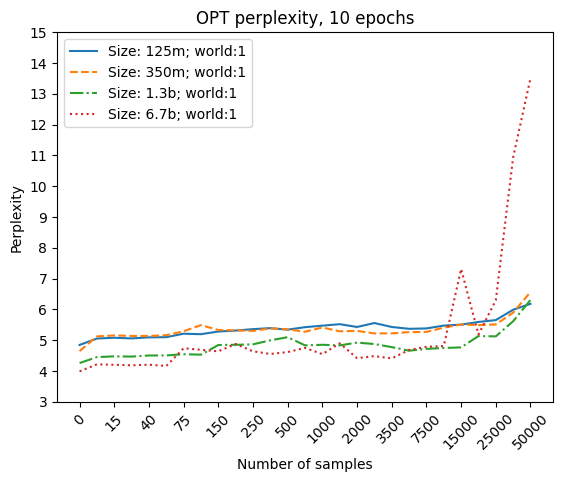}%
    \label{fig:fig3a}%
  }

    \centering
  \subfloat[Total-Variation Distance]{%
    \includegraphics[width=0.33\textwidth]{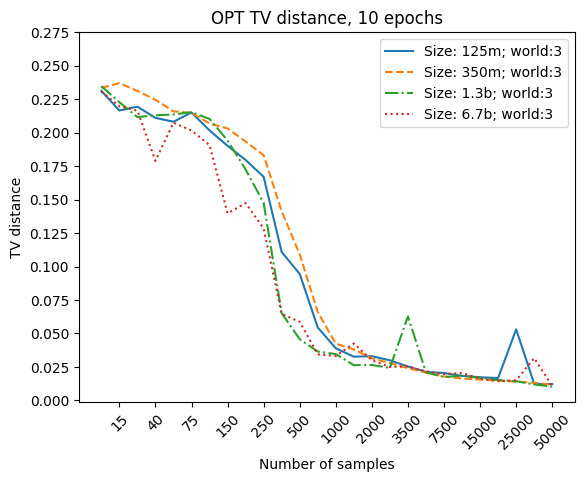}%
    \label{fig:fig1b}%
  }\hfill
  \subfloat[Structure Scores]{%
    \includegraphics[width=0.33\textwidth]{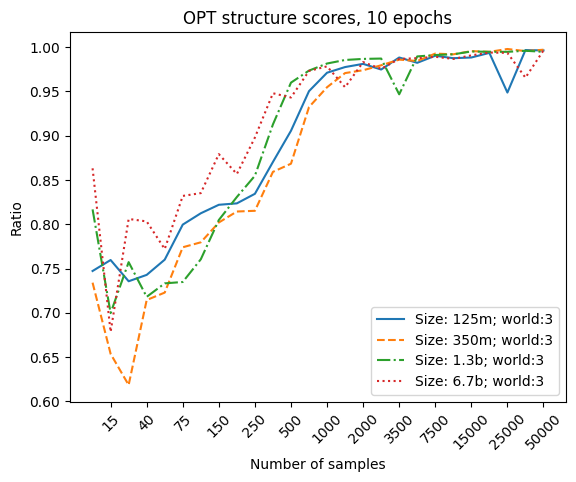}%
    \label{fig:fig2b}%
  }\hfill
  \subfloat[Perplexity]{%
    \includegraphics[width=0.33\textwidth]{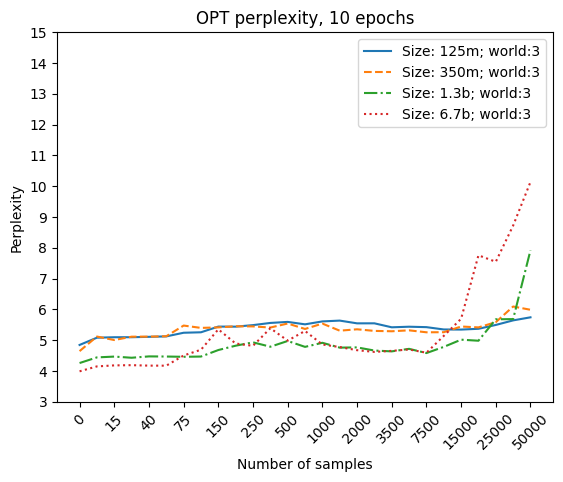}%
    \label{fig:fig3b}%
  }
  \caption{Results for OPT models. In the first row are the results for world-1 and in the second are the results for world-3. In cases, we plot the metric as a function of the number of training samples.}
  \label{fig:combined}
\end{figure*}

\begin{figure*}[t]
  \centering
  \subfloat[Total-Variation Distance]{%
    \includegraphics[width=0.4\textwidth]{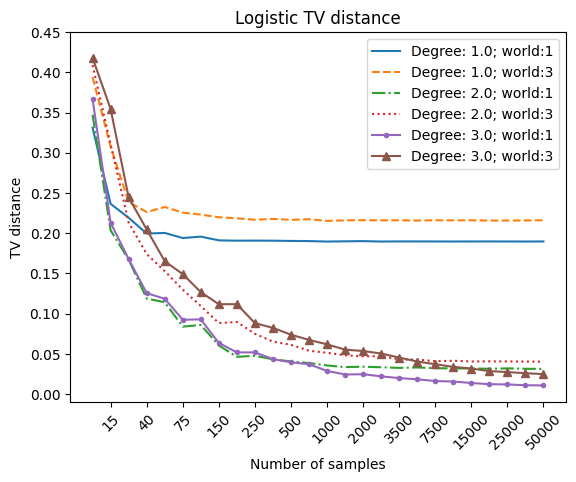}%
    \label{fig:fig1c}%
  }\hspace{1cm}
  \subfloat[Structure Scores]{%
    \includegraphics[width=0.4\textwidth]{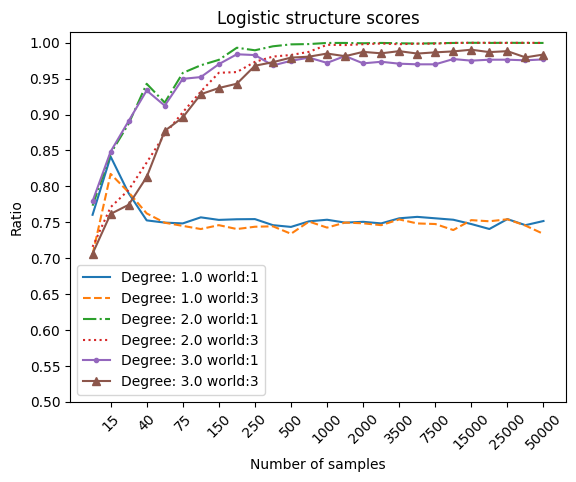}%
    \label{fig:fig2c}%
  }
  \caption{Results for Logistic Regression models.}
  \label{fig:combined2}
\end{figure*}

\begin{figure*}[t]
  \centering
  \subfloat[Total-Variation Distance]{%
    \includegraphics[width=0.4\textwidth]{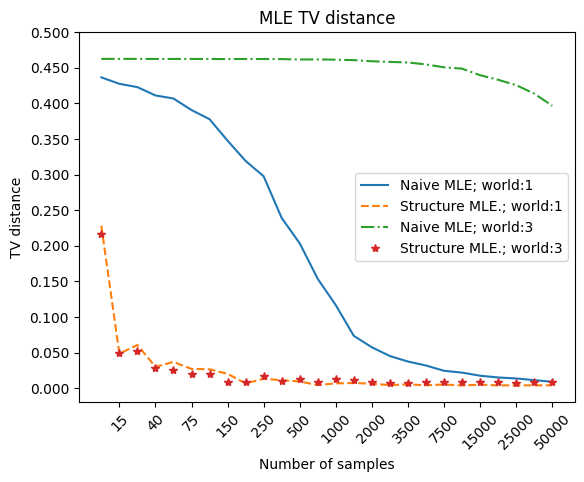}%
    \label{fig:fig1d}%
  }\hspace{1cm}
  \subfloat[Structure Scores]{%
    \includegraphics[width=0.4\textwidth]{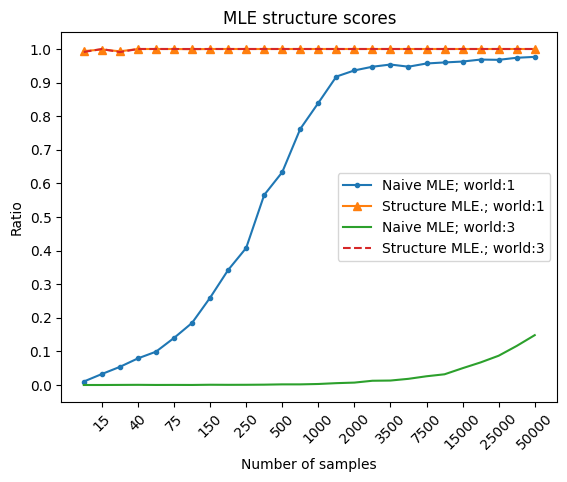}%
    \label{fig:fig2d}%
  }
  \caption{Results for MLE models.}
  \label{fig:combined3}
\end{figure*}

\begin{figure}[t]
\centering
\includegraphics[width=0.48\textwidth]{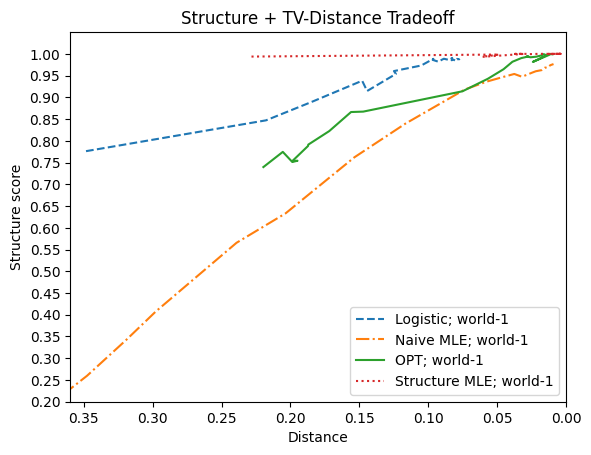}%
\caption{Tradeoff between TV-distance and Structure score. OPT represents the average distances and average scores for all $4$ sizes with $5$ seeds each.}
\label{fig: tradeoff}
\end{figure}

\paragraph{Models.}
We use the OPT models \cite{zhang2022opt} which come in various sizes. We used the 125m, 350m, 1.3b, and 6.7b parameter models. We used the released weights as initialization and then trained for the task. 

For training, we use Low-Rank Adaptation (LORA, \citet{hu2021lora}) with rank$=64$. 
We trained the models for 10 epochs with learning-rate $=2\cdot 10^{-4}$. We found that although more epochs improve results, the gain is not substantial for larger epoch numbers.

We found that training with randomly initialized model weights is significantly more challenging compared to a pretrained model. We also found LORA to be more stable, compared to finetuning a full model, and it also better preserves the perplexity of language tasks. Therefore we provide results with these settings only.

\subsection{Zero-shot Experiments}

We evaluated our learning tasks with off-the-shelf models in two zero-shot settings. In one setting we instructed the model to learn the distribution of the data as in-context prompts. In the second setting, we instructed the model to write functions that parse the data and then input the parsed values to a simple logistic regression model.

\paragraph{In-Context Learning.}
In this setting, the model is given a prompt with a list of observations and is asked to give the probability for some test observations.

Assuming a large number of samples is required for sufficient learning, this method is limited to models that are capable of large context windows, such as GPT4. Also, obtaining estimates for all possible inputs (which is required to compute the expected TV distance) is prohibitively expensive. Nevertheless, results for a few sample cases can still provide insight into the learning capabilities of this method.

We ran this setting for two random world settings, each with two random sets of training samples. We put in the context either 10, 100, or 1000 samples (with 1000 for the GPT4 models only). 
For the exact prompts see Appendix \ref{appendix: A1}.

\paragraph{Pipeline Learning.}
Inspired by the distinction between the physical rules and the latent distribution, we propose using LLMs as a step in a workflow pipeline. The model is given some background and examples and is asked to generate a parsing function that will be used with a Logistic Regression model (without polynomial features). 

We note that simply parsing the input as a list of values is insufficient for this task since the model does not consider multiplications. On the other hand, calculating the total torque is impossible since some variables are latent.

We tested $3$ OpenAI models: gpt-3.5-turbo-0125, gpt-4-turbo-2024-04-09, and gpt-4o-2024-05-13. We sampled $3$ worlds, each with $3$ different random sample sets, and generated prompts for the models. We also added different levels of hints to assist the model.
For more details about the prompts we used see Appendix \ref{appendix: A2} 

The models generate parse functions which we run on random test sets (using the same sets across all models). 
We then measure the average TV distance between the predictions and the real probabilities. In cases where the model returns an error, we mark the distance as 1.

\section{Results}

Here we report and plot the performance of the various models and methods.

We conducted the main experiments on two randomly chosen world settings. 
The first setting (World-1) has visible variables of \textit{mass1}, \textit{distance1}, \textit{side1}, \textit{mass2}, and \textit{side2}. The second setting (World-3)\footnote{World-1 and World-3 were generated with $\text{seed}=1,3$, respectively. We conducted the experiments with World-3 for variety, as World-2 was similar to World-1.} has visible variables of \textit{density1}, \textit{volume1}, \textit{mass1}, \textit{distance1}, \textit{side1}, \textit{density2}, \textit{volume2}, \textit{mass2}, and \textit{side2}. Notice that the mass variables are redundant in the second setting. 
In both settings, there were two objects and \textit{distance2} was latent. In World-1, the mean of the latent variable \textit{distance2} is $2.668$; in world-3, it is $3.203$. In both cases, the variance is $1$.

In Figure \ref{fig:combined} we report the TV distances and structure scores for the Transformer models. Additionally, we report the perplexity of the trained Transformers on a portion (first 500 documents) of Wikitext-2.\footnote{\url{https://huggingface.co/datasets/Salesforce/wikitext/viewer/wikitext-2-raw-v1}} These values serve as indicators of the extent to which the textual pretraining is affected by the task-specific finetuning.

In Figures \ref{fig:combined2} and \ref{fig:combined3} we report the TV distances and structure scores for the Logistic Regression and MLE models, respectively.

\paragraph{Zero-shot scores.}
Results for the pipeline learning experiment are reported in Table \ref{tab:zeroshot}. 

\begin{table}
    \centering
    \begin{tabular}{l|cc|cc}
        Model & {\small$TV_{pipe}$} & {\small $<0.1$} & {\small $TV_{ICL}$} & {\small $<0.1$}\\
        \hline \hline
        GPT-3.5 & 1. &  0. & 0.424 & 0.\\
        GPT-4 & 0.55 &  \textbf{0.22} & \textbf{0.207} & 0.\\
        GPT-4o & \textbf{0.51} & 0.037 & 0.235 & 0.\\
    \end{tabular}
    \caption{Results for the zero-shot experiments. $TV$ represents the average TV distance over all samples in all settings and $<0.1$ represents the ratio of experiments in which the (average) TV distance was smaller than 0.1. For TV lower is better and for $<0.1$ higher is better. Scores are reported for both the In-Context Learning (ICL) and pipeline methods.}
    \label{tab:zeroshot}
\end{table}

\section{Discussion}

As World-3 is clearly more complex than World-1, it is interesting to see how this affects different models. In Na\"{i}ve-MLE we see a significant performance gap between the settings, with the method only starting to learn with $\approx 10K$ samples in World-3. In Appendix \ref{appendix: theoretical} we derive simple bounds for the expected squared TV distance of this method and show that the extra variables have a substantial effect.
In Logistic Regression and OPT models, there is a consistent gap between the two settings but it is not substantial. In Structure-MLE the performance is practically the same.
This fits a general trend regarding the provided knowledge about the structure. The more the model is provided with structure, the less the effect of additional variables.

In Figure \ref{fig: tradeoff} we plot the tradeoff between the TV distance and the structure score (\ref{structure-score}). Since the Na\"{i}ve-MLE model makes no explicit assumptions regarding the structure, we can see its curve as representing the structure score that can be achieved without explicit learning.
This curve bounds the curves from below.
From above, the Structure-MLE bounds all other methods, as it is provided with full structure knowledge by definition. In between, we see that Logistic Regression has a higher structure score (per error) compared to OPT models. The general trend we see is that Transformer models seem to learn the structure to some extent, but it is not as strong as in models that are given stronger assumptions.

In most of our experiments, the models achieve low error. The exceptions are logistic regression with polynomial-degree 1 (for both worlds) and na\"{i}ve-MLE for world-3. In the first case, the model makes strong assumptions that simply do not fit the world. In the second case, the lack of assumptions regarding the world leads to extremely high sample complexity.

Among different settings of the Transformers, we find that larger models learn with fewer samples. However, smaller models seem to preserve the textual pretraining for longer learning.
Additional epochs improve the results, but only up to a certain point. 

The trend regarding the provided structure is aligned with the number of parameters in the models. Na\"{i}ve-MLE has $|states| ^{|values|}$ parameters. Structure MLE has 1 parameter. Logistic regression has $|states| ^{degree}$. In this respect, the OPT model is an exception, as larger models have better performance. 

While our zero-shot experiments generally show low results, they do show promising directions. With In-Context Learning, 
we find that LLMs, even strong long-context ones, struggle with this task. It seems then that the models implicitly apply simplistic heuristics instead of rigorous analysis.
In the pipeline experiment, we see that despite the poor performance, there is a clear hierarchy in which GPT4 and GPT4o clearly outperform GPT3.5. This shows that, to some extent, LLMs can be used as components in a pipeline that uses other models. This type of approach was described in \citet{wong2023word}, and we view it as a promising approach for the future.

This observation impacts many highly studied tasks that involve components that do not fall under the description of ``natural language'', such as chess \cite{toshniwal2022chess, feng2024chessgpt} and arithmetics \cite{yuan2023large}. Our findings suggest that perhaps tools that were designed for natural language are not optimal for these tasks. 

We note that although our experiments address finetuning and inference, the findings are relevant for pretraining too. Our findings show that the training data can contain information, that can be captured by simple models, but LLMs may not capture.

\section{Conclusion}

In this paper, we presented a novel framework for generating experimental worlds from a common physical setting, with easily manipulable distributions. The framework allows sampling from the ground-truth model and enables carefully controlled experiments in learning the distributions. We applied various methods, from classic learning algorithms to various sizes of Transformers. The methods range from highly structure-aware to structure-agnostic. 

Our findings show that even in a very simple physical setting, models that make stronger assumptions as to the structure present better sample complexity. Specifically, in the given setting, simple structure-based models like Logistic Regression and full structure MLE can be substantially more sample-efficient compared to Transformer models.
We further propose an approach to leverage LLMs as 
part of a pipeline that involves a classic learning algorithm. We show that models still struggle with this task but show a promising trend, as newer models show substantial improvements over older ones.

\section*{Limitations}
We stress that our experiments with different models differ in the information that is given to the model. For example, the baselines receive tabular data variables whereas the Transformer receives text. Consequently, the comparison is for analysis purposes and is not a strict comparison between methods.

We also note that although inspired by real-world physical settings, the data in our experiments is not distributed in anything like naturally occurring text. 

Regarding the zero-shot experiments, we note that the development of our prompt was based on worlds that were not used in the main experiments. However, the worlds are all similar to some extent so the generality of the results is limited.

\section*{Acknowledgements}
The authors thank Moshe Friedman, Yuli Slavutsky, and Eran Malach for their valuable insights.
This research was supported by grants from the Israeli Ministry of Science and Technology, the Israeli Council for Higher Education, and the Israel Science Foundation (grant no. 2424/21).

\bibliography{anthology,custom}
\bibliographystyle{acl_natbib}

\appendix
\section{Prompts For GPT4}
\subsection{In-Context Learning} \label{appendix: A1}


For In-Context Learning, we used the prompt:

\begin{quote}
    Assume we have a model representing a lever on a fulcrum, with two objects on it. The first object is on the right and the second is on the left.

    I'll give you a list of partial observations of the states of the model. Notice that some values might be latent. Then I'll ask you to give me the probability for the continuation of some prompt, based on the distribution you can derive from the samples. Be prompt in your answer.
    
    Samples:
    <list of samples>
    
    Question:
    I'll give you a list of prompts. Give me a python list with the probabilities of "L", one probability for each input.
    Samples:
    <list of test samples>
    Give me a list only with no additional explanations.
\end{quote}

\subsection{Model Recommendation} \label{appendix: A2}
Asking the model for a recommended learning method, we used the prompt:

\begin{quote}
    Assume we have a model representing physical setting.
    <add the first hint here>
    Here's a list of partial observations of the states of the model. 
    <add the second hint here>
    <add the third hint here>
    Samples:
    <list of samples>
    
    I want to learn the distribution using Statistics or Machine Learning. 
    Specifically, I want to use Logistic Regression to predict the balance probabilities of new samples. Here is an example of the code:
    
    \begin{small}
    \begin{verbatim}
    ```python
    def fit_lr(X, y):
        from sklearn.linear_model \
            import LogisticRegression
        model = LogisticRegression(
            max_iter=10000, 
            solver='saga')
        model.fit(X, y)
        return model
        
    def predict_lr(model, X):
        return model.predict_proba(X)
    ```
    \end{verbatim}
    \end{small}
    
    Write me python function 
    parse\_samples$()$, that parses each sample and creates a feature function that can be used in the snippet above.
    Make sure the function is appropriate for both training and inference.
    Give me code only.
\end{quote}

The hints that were (possibly) provided were: 
\begin{quote}

(1) We have a lever on a fulcrum with objects on the lever. \\
(2) Notice that some variables might be latent. \\
(3) Notice that the distance of the last object is latent. \\
                 
\end{quote}

\section{Theoretical Sample Complexities} \label{appendix: theoretical}

Here we provide theoretical analysis for the sample complexity of the Naive-MLE baseline. For the loss function, we consider the expected squared TV distance.

The Naive-MLE estimates an independent distribution for $y$ for each set of values of the other variables, $x$. 
For each assignment $x$, assume we have $N_{x}$ samples, and estimator is $\hat{p}(y=1|x) = \frac{1}{N_{x}} \cdot \mathbbm{1}_{y=1}$.
As a Bernoulli random variable, we know that the variance of $\hat{p}$ is $E[(\hat{p}-p)^2]=\frac{p(1-p)}{N_{x}}$.
This can be bounded by $\frac{1}{4N_{x}}$.

We have $$E[TV^2(\hat{p}, p)]=\frac{1}{4} \cdot E[(\hat{p}-p)^2] \leq \frac{1}{16N_{x}}$$ 
So, for any $\epsilon^2 > 0$, if we have $N_{x} \geq \frac{1}{16\epsilon^2}$ samples then the expected squared error will be $\leq \epsilon^2$.

Now, we need to bound the probability of $N_{x} < \lceil \frac{1}{16\epsilon^2} \rceil \leq N^*$, given a total number of samples $N$. 
$N_x$ is distributed as a Binomial random variable with parameters $n=N, p_x=p(X=x)$. 
Following \citet{arratia1989tutorial}, we can use the tail bound 
\begin{multline}
    Pr( N_x \leq N^*) \leq \exp \big(-N D(\frac{N^*}{N}||p_x)\big)
\end{multline} \label{bound}
for $N^* \leq Np_x$.

Since we assumed i.i.d. for the observed inputs, we can use an identical bound for each case. In the simple case, with $3$ input variables with $5$ values each (similar to World-1 when combining the distance with the side). 
For a squared error of less than $\epsilon^2 = 0.05^2$ and get $N^* \geq 25$.

Taking $N^* = 32$, the upper bound in \ref{bound} together with the union bound (for $125$ options) gives probability $p > 1-0.0092$ for all inputs to have at least $25$ samples. This choice for $N^*$ yields $N = 32 \cdot 125 = 4000 $ which is similar to the empirical results we got.

This same bound for a case with $8$ variables (similar to World-3) this bound goes up to $N = 32 \cdot 5^8 = 1.25 \cdot 10^7$ samples.

\end{document}